%% file: main.tex
\documentclass{article}

\usepackage[preprint]{neurips_2024}

\usepackage{graphicx}
\usepackage{amsthm}
\usepackage{amsmath}
\usepackage{amssymb}
\usepackage{amsmath,amssymb,amsfonts}
\usepackage{amsmath}
\usepackage{amssymb}
\usepackage{float} 
\usepackage{multirow}
\usepackage{subcaption}
\usepackage{graphicx}
\usepackage[table]{xcolor}
\usepackage{colortbl}
\usepackage{xcolor}
\usepackage{tabularx}
\usepackage{makecell}
\usepackage{booktabs}

\usepackage{enumitem}

\usepackage{wrapfig}

\usepackage{graphicx}
\usepackage{subcaption}

\usepackage[utf8]{inputenc} % allow utf-8 input
\usepackage[T1]{fontenc}    % use 8-bit T1 fonts
\usepackage{hyperref}       % hyperlinks
\usepackage{url}            % simple URL typesetting
\usepackage{booktabs}       % professional-quality tables
\usepackage{amsfonts}       % blackboard math symbols
\usepackage{nicefrac}       % compact symbols for 1/2, etc.
\usepackage{microtype}      % microtypography
\usepackage{xcolor}         % colors

\usepackage[ruled,vlined]{algorithm2e}
\usepackage{amsmath}
\usepackage{float}

\usepackage{graphicx}
\usepackage{subcaption}
\usepackage{adjustbox}

\usepackage{amsmath}

\let\subparagraph\paragraph
\usepackage[compact]{titlesec}

\titlespacing{\section}{0pt}{1ex}{1ex}
\titlespacing{\subsection}{0pt}{1ex}{0.5ex}
\titlespacing{\subsubsection}{0pt}{0.8ex}{0.5ex}

\DeclareMathSizes{9}{8.2}{5}{3}
\newcommand{\sectionfontsize}{\fontsize{13pt}{15pt}\selectfont}
\newcommand{\subsectionfontsize}{\fontsize{12pt}{14pt}\selectfont}
\titleformat{\section}
  {\sectionfontsize\bfseries}
  {\thesection}{1em}{}

\titleformat{\subsection}
  {\subsectionfontsize\bfseries}
  {\thesubsection}{1em}{}

\title{LExI: \underline{\text{L}}ayer-Adaptive Active \underline{\text{Ex}}perts for \\ Efficient MoE Model \underline{\text{I}}nference}

\author{
  Krishna Teja Chitty-Venkata, Sandeep Madireddy, Murali Emani, Venkatram Vishwanath\\
  \texttt{ schittyvenkata@anl.gov, smadireddy@anl.gov, memani@anl.gov, venkat@anl.gov  @anl.gov} \\
  Argonne National Laboratory, Lemont, IL 60439, USA
}

\begin{document}

\maketitle

\begin{abstract}

% Mixture-of-Experts (MoE) models offer a scalable alternative to dense architectures by activating a subset of experts per token. 

% Earlier post-training optimizations, such as Inter and Intra expert pruning in MoEs, reduce memory but do not enhance the compute performance.  

% The current designs uniformly use a fixed number of active experts across the network, leading to unnecessary computational overhead. We first show that pruning only enhances memory but not compute performance on GPU and propose a better alternative to pruning. We propose LExI, a data-free optimization strategy to assign an optimal number of active experts per layer in a pretrained MoE model. Our method leverages only the model weights to estimate the layer importance to assign the number of active experts. Experimental results across multiple state-of-the-art language and vision MoE benchmarks show that our approach reduces compute performance during inference with negligible model accuracy loss, offering an efficient MoE inference solution compared to pruning weights and experts. The code will be released in the future. 

Mixture-of-Experts (MoE) models scale efficiently by activating only a subset of experts per token, offering a computationally sparse alternative to dense architectures. While prior post-training optimizations, such as inter- and intra-expert pruning, reduce memory usage but provide limited gains in inference-time compute efficiency. Moreover, existing MoE architectures typically activate a fixed number of experts uniformly across all layers, resulting in redundant computation and suboptimal performance. In this work, we first demonstrate that MoE pruning strategies improve only the memory footprint but do not significantly improve inference performance on GPU using optimized frameworks such as vLLM. To address this, we introduce \textbf{LExI}, a data-free optimization technique that determines the optimal number of active experts per layer in a pretrained MoE model. LExI leverages only the model’s weights to estimate the relative importance of each layer and adaptively assigns the number of active experts accordingly per layer. Experiments on state-of-the-art language and vision MoE benchmarks demonstrate that LExI significantly outperforms traditional MoE pruning approaches in terms of inference efficiency with negligible accuracy loss. For example, using LExI, Qwen1.5-MoE achieves the same throughput on Nvidia H100 GPU with 10\% better accuracy than traditional expert pruning. %The code will be released publicly.

% \SM{add quantitative numbers}

% compute performance \SM{this sentence is not clear} 
\end{abstract}

% \section{Introduction}

\input{Sections/1_Introduction}

\input{Sections/2_Related_work}

\input{Sections/3_profiling_latency}

\input{Sections/4_Expert_pruning_varying_topk}

\input{Sections/5_results}

\input{Sections/6_abalation}

\input{Sections/7_conclusion}

\section*{Acknowledgements}
This research used resources of the Argonne Leadership Computing Facility, a U.S. Department of Energy (DOE) Office of Science user facility at Argonne National Laboratory and is based on research supported by the U.S. DOE Office of Science-Advanced Scientific Computing Research Program, under Contract No. DE-AC02-06CH11357

\bibliography{main.bib}
\bibliographystyle{plainnat}

\appendix
\input{Sections/8_Appendix}

\end{document}

%% file: Sections/1_Introduction.tex
\section{Introduction}

% A Mixture of Expert (MoE) model is a powerful architecture to enable efficient inference by dynamically activating only a subset of model parameters for each token. MoEs route each token to a small number of expert networks instead of processing the input tokens through all the network parameters, significantly reducing computational cost while maintaining or improving model capacity.
% Training MoEs typically involves jointly learning the expert weights and a routing/gate function. Auxiliary loss, such as load balancing, is commonly added during training to ensure balanced training and avoid expert collapse. Prominent examples of MoE-based models include Large Language Models (LLMs) like Mixtral (\cite{jiang2024mixtral}), Phi-3.5-MoE (\cite{abdin2024phi}), Qwen1.5-MoE-A2.7B (\cite{qwen_moe}), OLMoE (\cite{muennighoff2024olmoe}), Jamba (\cite{lieber2024jamba}), as well as Vision Language models (VLMs) like MolmoE (\cite{deitke2024molmo}) and DeepSeek-VL2 (\cite{wu2024deepseek}), which leverage sparse expert activation to handle high-dimensional multimodal inputs efficiently. While this sparse activation reduces computation compared to dense models, MoE models remain computationally expensive at inference time, particularly due to the uniform use of a fixed $k$ across all layers and tokens.

\begin{figure*}
    \centering
     \includegraphics[width=0.95\linewidth]{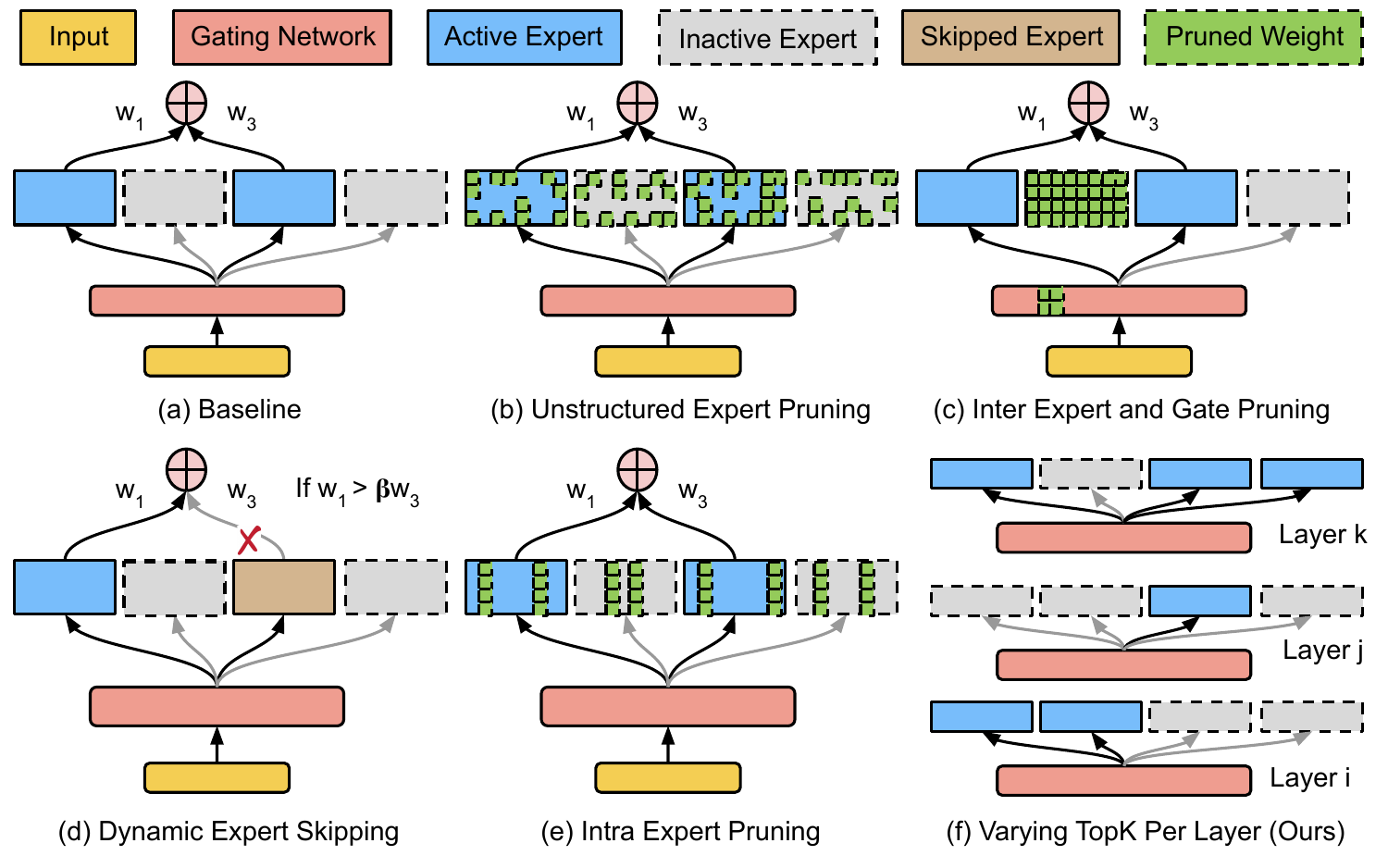}
        \caption{Overview of MoE Optimization Methods. (a) Baseline Trained Model (b) Unstructured Expert Pruning (SparseGPT (\cite{frantar2023sparsegpt}), Wanda (\cite{sun2023simple}), MoE-Pruner (\cite{xie2024moe})) (c) Inter Expert Pruning (NAEE (\cite{lu2024not})) (d) Dynamic Expert Skipping (NAEE (\cite{lu2024not})) (e) Intra Expert Pruning (MoE-I$^{2}$ (\cite{yang2024moe})) (f) \textbf{LExI}: Static Varying Active Experts (Topk) Per Layer (Ours)}   
        \vspace{-4mm}
        \label{fig:paged_compression}
    \captionsetup{justification=centering}
\end{figure*}
%\vspace{3mm}

% \SM{Include a figure in page 1 or 2 that shows the main take away of the work}

Large Language Models (LLMs) and Vision Language Models (VLMs) have achieved remarkable performance through model scaling, but require tremendous compute and memory resources. Mixture-of-Experts (MoE) models have emerged as a promising approach to increase model capacity without a proportional rise in inference cost. In an MoE, multiple expert subnetworks are trained, and a sparse gating or router network activates only a small subset of experts ($k$ experts) per input token. This sparse computation allows MoEs to outperform dense models with the same number of active model parameters.  Prominent examples of MoE-based LLM include Mixtral (\cite{jiang2024mixtral}), Qwen1.5-MoE-A2.7B (\cite{qwen_moe}), OLMoE (\cite{muennighoff2024olmoe}), and Vision Language Models (VLM) such as MolmoE (\cite{deitke2024molmo}) and DeepSeek-VL2 (\cite{wu2024deepseek}). 

One commonly used post-training optimization strategy for MoE models is expert pruning, which removes redundant experts. Recent methods such as NAEE (\cite{lu2024not}), MoE-Pruner (\cite{xie2024moe}) EEP (\cite{liu2024efficient}) and MoE-I$^{2}$ (\cite{yang2024moe}) introduce various pruning strategies. For example, NAEE identifies and removes entire experts from a pretrained MoE model, while MoE-I$^{2}$ prunes the inner dimensions within each expert's MLP. Although these methods reduce the memory footprint, our performance evaluation across several state-of-the-art  MoE models using the widely adopted vLLM inference framework reveals a critical limitation: \textit{pruning does not consistently translate into faster inference}, and in some cases, it even degrades performance. This degradation is primarily due to the sparse structure of the MoE models itself. 
% \SM{why, and reference?}. This is our observation and I written why this is happening below.   
In expert pruning, the input token still needs to be routed to the same number of top-$k$ experts, as determined by the router. While some experts are pruned, the remaining ones must process a disproportionately larger number of tokens, increasing their computational load. In the batched inference scenario, this load imbalance can lead to longer processing time per expert, thereby increasing overall latency. While aggressive expert pruning levels
% \SM{of experts?} 
can yield noticeable speedups, they typically result in significant accuracy degradation, making them impractical. Moreover, these expert pruning approaches usually rely on training data for pruning experts. 
Current MoE architectures employ a fixed top-$k$ routing mechanism, where the same number of experts are activated for each input token across all layers. This static design is suboptimal as different layers may require varying levels of expert capacity depending on computational needs (\cite{guo2024dynamic}). 
% \SM{add citation}. %not require the same degree of expert capacity. 
Beyond computation, inter-GPU communication overhead also becomes a significant performance bottleneck in MoEs. Increasing the number of active experts per token increases the volume of communication operations such as all-reduce and broadcast, further adding to the inference cost. To address some of these limitations, NAEE (\cite{lu2024not}) proposed a token-aware dynamic expert skipping strategy, which selectively skips an expert during inference. However, this strategy is highly tailored to the dataset and cannot work beyond top-$k$=2. In summary, existing MoE optimization strategies often rely on calibration sets for pruning or routing adjustments, making them unsuitable in deployment settings where access to data or retraining is infeasible. In addition, the dataset-driven solutions will make pruned models more optimized to that calibration dataset, potentially degrading the performance in unseen settings.

Motivated by these insights, we propose \textbf{\textit{LExI}}, a novel post-training layer-adaptive active expert allocation mechanism that determines the optimal number of active experts per layer without depending on any dataset. Our method relies on the key observation that not all layers contribute equally to the final model performance and that expert redundancy varies significantly across depth. This raises a fundamental question: \emph{Can we reduce the number of active experts per layer irrespective of the input token without sacrificing accuracy?} In particular, is it possible to statically assign different top-$k$ values to each layer, so that every layer uses just enough experts to retain its contribution, while improving overall inference efficiency? By making layer-adaptive expert allocation decisions, LExI reduces computational overhead across the model irrespective of the input token, offering a more efficient alternative to the traditional fixed top-K routing. 
% In this paper, we demonstrate that LExI substantially improves the efficiency of MoE inference with minimal accuracy degradation. 
Our experiments show that LExI outperforms existing expert pruning techniques in both task performance and runtime efficiency. By reducing the average number of activated experts per layer, LExI reduces latency and memory bandwidth usage while maintaining competitive task accuracy across diverse tasks.

\paragraph{Contributions.} Our key contributions are as follows:
\begin{itemize}[noitemsep,topsep=0pt,leftmargin=*]
    \item We introduce \textbf{LExI}, a novel dataset-free optimization 
    % \SM{is this a standard term?} yeah 
    technique for static active expert assignment in pretrained MoE models. LExI is simple to implement and serves as an efficient, plug-and-play solution for inference across various frameworks.
    
    \item We propose a data-free profiling strategy to estimate the sensitivity of each expert using only model expert weights. LExI  combines this one-time profiling with evolutionary search to determine the optimal active experts layer in a computationally efficient manner.
    
    \item Unlike prior methods that demonstrate improvements on a narrow set of MoE models (Mixtral-8x7B), our approach generalizes across multiple state-of-the-art MoE architectures in both language and vision domains.
    
    \item We empirically show that expert pruning in MoE models does not significantly improve inference performance, and in some cases, can degrade it due to architectural sparsity and load imbalance. Our method provides a viable alternative to pruning, improving both accuracy and hardware efficiency without requiring retraining or access to calibration data.
\end{itemize}

%% file: Sections/2_Related_work.tex
\section{Background and Related Work}

\noindent\textbf{Mixture of Experts.} Mixture-of-Experts (MoE) architectures improve the scalability and efficiency of LLMs/VLMs by introducing sub networks or experts. For a given input $x$, the output $y$ of an MoE module is computed as a weighted sum over all the active top-$k$ experts: $y = \sum_{i=1}^{top-k} G(x)_i \cdot E_i(x)$, where $\quad G(x) := \text{Softmax}(\text{TopK}[x \cdot W_g])$. The $\text{TopK}[\cdot]$ function selects the top-$k$ experts with the highest gating scores, and Softmax normalizes their scores into a probability distribution. $G(x) \in \mathbb{R}^N$ is the gating vector representing the importance weights assigned to each expert in the top-$k$ selected ones, and $E_i(x)$ denotes the output of the $i$-th expert given input $x$. Each expert $E_i$ is typically an FFN and constitutes the dominant portion of the parameters model (e.g., up to 96\% in Mixtral).

\noindent\textbf{Pruning Large Language Models.} Model pruning is a well-established technique to reduce inference costs by removing less important parameters. Recent works such as SparseGPT (\cite{frantar2023sparsegpt}) and Wanda (\cite{sun2023simple}) demonstrated one-shot pruning methods that introduce unstructured or semi-structured sparsity in weight matrices, cutting up to 50\% of parameters in GPT-scale models with minimal perplexity loss. These approaches solve layer-wise reconstruction or use weight magnitude heuristics to remove weights, and have been applied to models as large as GPT-175B. However, the irregular weight sparsity pattern they induce often requires specialized hardware support to realize actual runtime speedups (\cite{zhou2021learning}), and may suffer degraded efficiency on general-purpose accelerators.

\noindent\textbf{Expert Pruning and Compression in MoE Models. } 
% MoE architecture can enable an extreme form of structured sparsity, by removing entire experts. 
Since MoE models typically allocate the vast majority of their parameters to the expert sub-networks, pruning even a subset of experts can lead to substantial memory savings. NAEE (\cite{lu2024not}) is a post-training expert pruning framework to permanently remove unimportant experts without needing to re-train the model. %Lu et al. show that by measuring each expert’s contribution to model output on a small calibration set, one can identify and prune away the least significant experts, greatly reducing model size and memory footprint. 
By evaluating each expert's contribution to the model's output on a small calibration set, NAEE identifies and permanently prunes the least significant experts.
Furthermore, NAEE also introduced an inference-time policy to dynamically skip experts for certain tokens on the fly, effectively adjusting the active expert count based on the input token. Another recent approach is MoE-I2 (\cite{yang2024moe}), which introduces a two-stage compression pipeline tailored for MoEs. In the inter-expert pruning stage, MoE-I2 performs a layer-wise analysis to prune a fraction of experts to prune. The authors also introduce intra-expert compression to reduce the inner dimensionality of an expert's FFN.  These advances in MoE-specific pruning highlight the growing interest in expert-level model trimming. Our proposed LExI method shares the overarching goal of exploiting expert redundancy to improve efficiency. However, rather than relying on static pruning, it focuses on adaptive expert utilization at inference time, offering a flexible and data-free alternative that preserves task performance while reducing computational cost.

%% file: Sections/3_profiling_latency.tex
\section{Experimental Setup and MoE Expert latency Profiling}

In this section, we evaluate the hardware performance of the MoE benchmarks to motivate our varying top-$k$ solution. We first provide a detailed experimental setup used in all our evaluations.

% \paragraph{Experimental Setup. } 
\noindent\textbf{Mixture-of-Experts Benchmarks.} We evaluate our method across a diverse set of MoE models spanning both language and vision-language domains. For LLMs, we consider \textit{Mixtral-8x7B-Instruct} (\cite{jiang2024mixtral}), \textit{Qwen1.5-MoE-A2.7B-Chat} (\cite{qwen_moe}), \textit{OLMoE-1B-7B-0924-Instruct} (\cite{muennighoff2024olmoe}), \textit{MiniCPM-MoE-8x2B} (\cite{hu2024minicpm}), and \textit{DeepSeek-V2-Lite-Chat} (\cite{liu2024deepseek}). For VLMs, we use \textit{DeepSeekVL2-Tiny} model (\cite{wu2024deepseek}). These models exhibit a wide range of MoE architectures, varying number of experts and active experts per token, enabling a robust evaluation of our proposed method across heterogeneous settings.

\noindent\textbf{Evaluation Benchmarks.} We benchmark LLMs on nine widely adopted language understanding tasks from the \texttt{lm-eval} (\cite{eval-harness}) suite: \textit{ARC-c} (\cite{Clark2018ThinkYH}), \textit{ARC-e} (\cite{Clark2018ThinkYH}), \textit{BoolQ} (\cite{clark2019boolq}), \textit{HellaSwag} (\cite{zellers2019hellaswag}), \textit{MMLU} (\cite{hendryckstest2021}), \textit{OpenBookQA} (\cite{OpenBookQA2018}), \textit{RTE} (\cite{wang-etal-2018-glue}), \textit{WinoGrande} (\cite{sakaguchi2019winogrande}). We report average accuracy across all these tasks to assess general-purpose language reasoning. For long content understanding, we employ \textit{Qasper} dataset (\cite{dasigi2021dataset}) from the LongBench suite (\cite{bai2024longbench}) and report the \textit{F1 score} as the evaluation metric. Additionally, we include a \textit{passkey retrieval} task (\cite{peng2023yarn}), where the accuracy is measured as the percentage of instances (over 100 iterations and varying depths) in which the model correctly identifies a passkey from the garbage context. To evaluate language modeling quality, we compute \textit{perplexity} on the \textit{C4} (\cite{dodge2021documenting}), \textit{PTB} (\cite{marcus1993building}), and \textit{WikiText-103} (\cite{merity2016pointer}) datasets. For vision-language models, we use three benchmarks from the VLMEvalKit (\cite{duan2024vlmevalkit}) suite: \textit{MME} (\cite{yin2023survey}), MMMU \cite{yue2024mmmu}, and \textit{ScienceQA} (\cite{lu2022learn}). These datasets span a broad spectrum of multimodal reasoning tasks and allow us to evaluate our method's effectiveness in VLM setting.

\noindent\textbf{Hardware and Software Setup.} All inference performance evaluations are conducted on \textit{NVIDIA H100 GPUs} with 80GB of HBM memory per GPU, supporting Tensor Cores for optimized matrix operations. We use \texttt{vLLM} (\cite{kwon2023efficient}) as our inference engine which is a high-performance framework with native support for MoE models via \textit{FusedMoE}, which fuses expert computation and routing to improve efficiency. Unless otherwise specified, all LLMs are deployed on 4 GPUs, while DeepSeek-V2-Lite-Chat and DeepSeekVL2-Tiny use 2 GPUs. We employ tensor parallelism across devices for all models. During inference, we use a batch size of 16, with input and output sequence lengths varied across models to comply with each model's maximum context length constraints.
% \noindent\textbf{Performance Metrics.} 
We report \textit{throughput} as our primary hardware performance metric, defined as the total number of tokens (input + output) processed per second. To calculate this, we first measure the \textit{end-to-end latency}, defined as the time elapsed from input prompt submission to the generation of the final output token, and then convert this latency into throughput. The metric for VLMs is the number of input (image + text) samples process per second.

% \SM{Can this be merged with any other section? this is too small for a section} KT:done

\noindent\textbf{Inter and Intra Expert Pruning.} Inter-pruning (\cite{lu2024not}) removes entire experts and their routing weights, reducing memory footprint while maintaining the same number of active experts per token during inference. Intra-pruning (\cite{yang2024moe}) targets the inner dimensions (FFN intermediate size) within each expert, preserving the expert count while reducing individual expert complexity. In our evaluation, we consider the following percentages of these pruning: \{12.5\%, 25\%, 50\%\}. 12.5\% inter pruning removes $1/8$th of the experts in each layer, whereas 25\% intra pruning prunes $1/4$th of the FFN dimension in each expert of each layer. The top-$k$ search space in our paper includes every integer from 1 up to the baseline pretrained top-$k$: ${1, 2, \ldots, \text{top-}k_{\text{baseline}}}$.

\begin{figure*}
\centering
    % Full-width legend at the top
    \includegraphics[width=\textwidth]{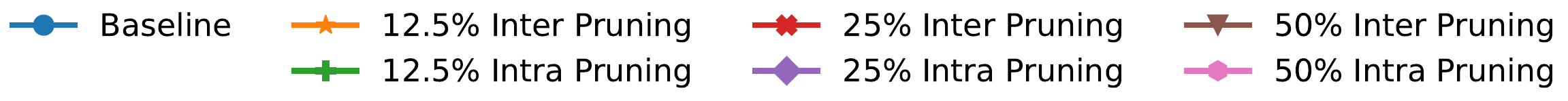}\\[0.2ex]
    % First row of three plots
    \begin{subfigure}[b]{0.32\textwidth}
        \centering
        \includegraphics[width=\linewidth]{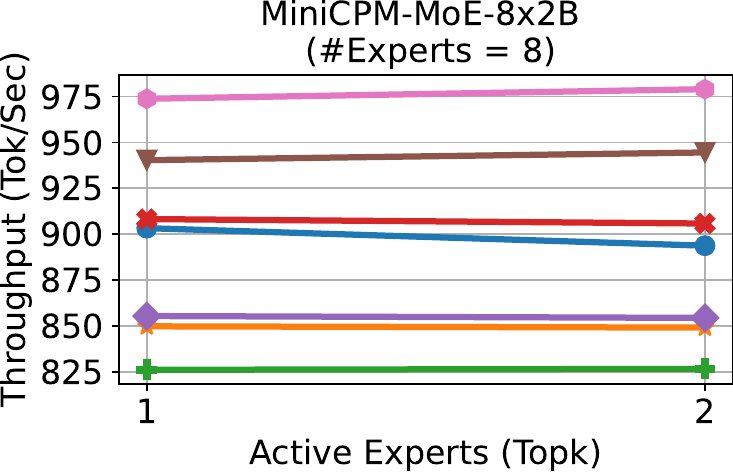}
        \caption{MiniCPM-MoE-8x2}  % (a) optional sub-caption
    \end{subfigure}\hfill
    \begin{subfigure}[b]{0.32\textwidth}
        \centering
        \includegraphics[width=\linewidth]{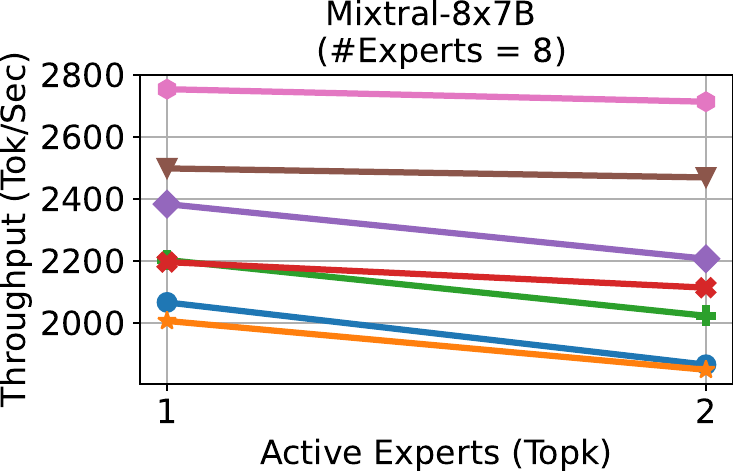}
        \caption{Mixtral-8x7B}  % (b) optional sub-caption
    \end{subfigure}\hfill
    \begin{subfigure}[b]{0.32\textwidth}
        \centering
        \includegraphics[width=\linewidth]{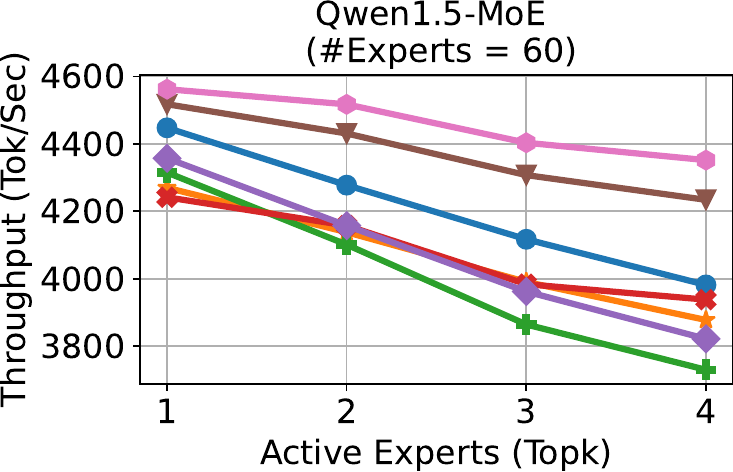}
        \caption{Qwen1.5-MoE-A2.7B}  % (c) optional sub-caption
    \end{subfigure}\\[0.2ex]
    % Second row of three plots
    \begin{subfigure}[b]{0.32\textwidth}
        \centering
        \includegraphics[width=\linewidth]{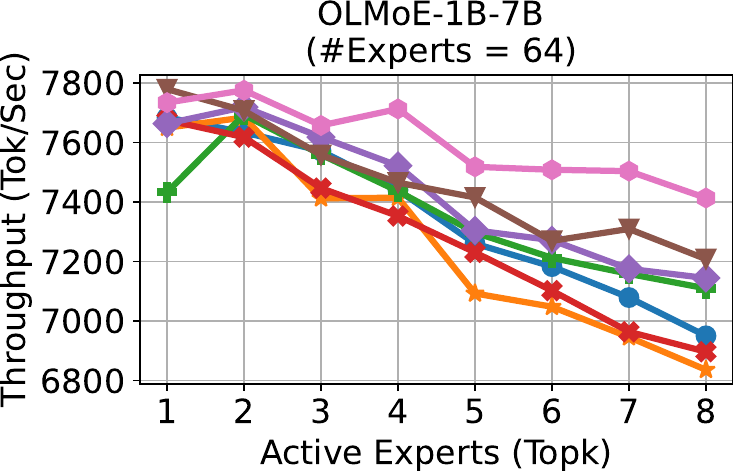}
        \caption{OlmoE-1b-7b-0924}  % (d) optional sub-caption
    \end{subfigure}\hfill
    \begin{subfigure}[b]{0.32\textwidth}
        \centering
        \includegraphics[width=\linewidth]{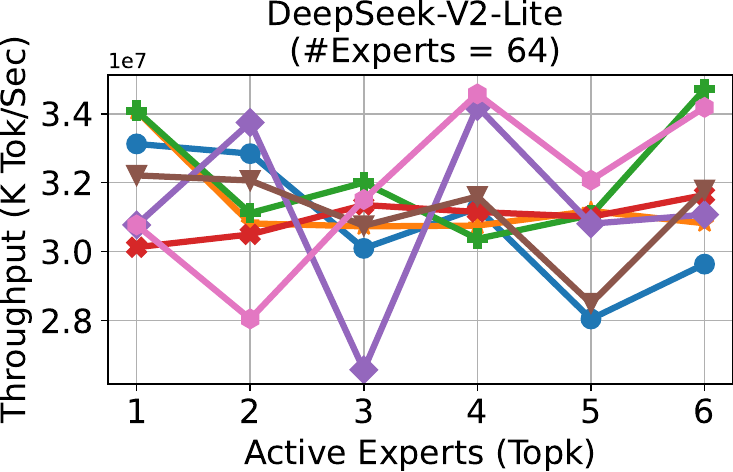}
        \caption{DeepSeek-V2-Lite}  % (e) optional sub-caption
    \end{subfigure}\hfill
    \begin{subfigure}[b]{0.32\textwidth}
        \centering
        \includegraphics[width=\linewidth]{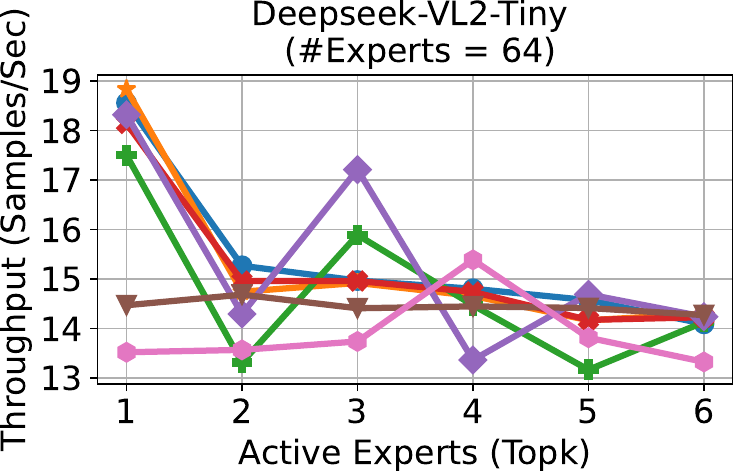}
        \caption{DeepSeekVL2-tiny}  % (f) optional sub-caption
    \end{subfigure}
    \caption{Throughput vs. Active Experts under Inter and Intra Expert Pruning}

    \label{fig:throughput-vs-topk}
\end{figure*}

% \noindent\textbf{Observations. } 
Figure \ref{fig:throughput-vs-topk} illustrates the throughput of the six benchmark MoEs under varying degrees of pruning and top-$k$. Models with fewer active experts (e.g., MiniCPM, Mixtral) show marginal gains with aggressive pruning, while models with more active experts (e.g., Qwen, OLMoE) exhibit complex interactions where pruning can improve or degrade throughput depending on token-to-expert routing balance and compute saturation. Notably, DeepSeek-VL2-Tiny shows throughput instability under pruning, suggesting higher sensitivity to expert load balance. The performance degradation stems from load imbalance across experts, leading to an increased number of tokens processed by each active expert.

% \paragraph{Why MoE Pruning Hurts performance?}

% \begin{algorithm}[H]
% \caption{Compute Output Frobenius Norm w.r.t. Baseline MoE}
% \label{alg:output_norm}
% \begin{algorithmic}[1]
% \Require Baseline MoE module $M_0$, pruned modules $\{M_p\}$, top-$k$ list $\mathcal{K}$, input shape $(B, L, H)$, iterations $N$
% \State Initialize: $\mathcal{N}[k][p] \gets [\,] \quad \forall k \in \mathcal{K}, \forall p \in \{M_p\}$
% \For{$i = 1$ to $N$}
%     \State Sample input: $x \sim \mathcal{N}(0, 1)^{B \times L \times H}$
%     \State Compute baseline: $M_0.\text{set\_topk}(k_0)$, $y_0 \gets M_0(x)$
%     \For{each $k \in \mathcal{K}$}
%         \For{each pruned module $M_p$}
%             \State $M_p.\text{set\_topk}(k)$, $y_p \gets M_p(x)$
%             \State $\mathcal{N}[k][p].\text{append}(\| y_p - y_0 \|_F)$
%         \EndFor
%     \EndFor
% \EndFor
% \For{each $k \in \mathcal{K}$ and $p \in \{M_p\}$}
%     \State $\mathcal{N}[k][p] \gets \text{mean}(\mathcal{N}[k][p])$
% \EndFor
% \State \Return $\mathcal{N}$
% \end{algorithmic}
% \end{algorithm}

%% file: Sections/4_Expert_pruning_varying_topk.tex
\section{LExI}

LExI implements a two-stage pipeline to determine the optimal number of active experts per layer.  In the first stage, it performs a one-time profiling to assess each layer's sensitivity to different top-$k$ values.
The sensitivity profiling methodology extends beyond expert allocation, serving as a foundation for diverse optimization problems such as layer-specific mixed-precision quantization or layer-wise pruning. 
% \SM{you said diverse optimization problems but mentioned only one}. 
The second stage employs a low-cost evolutionary search algorithm leveraging these sensitivity values as efficient proxies to identify the best performing top-$k$ for each layer.
% \SM{instead of optimal, say best performing?} 

% Algorithm \ref{alg:topk_perturbation} describes our Monte Carlo approach for quantifying MoE module robustness for each top-$k$ in the target search space. For each layer, we sample a random tensor from uniform distribution $\mathbf{X} \sim \mathcal{N}(0, 1)^{B \times L \times H}$. We first calculate the unperturbed output of the layer on this randomly generated input tensor followed by calculating the outputs for each top-$k$ in the search space. We measure the output deviation using the Frobenius norm between unperturbed output and the output generated by the target top-$k$ active experts. We repeat this process across several millions of iterations to capture the average perturbation for each candidate top-$k$ value. This approach effectively quantifies the sensitivity of the MoE routing mechanism to diverse set of input tokens, providing a principled metric for understanding how expert selection patterns affect each layer behavior. The method works because the Frobenius norm precisely measures the magnitude of output deviations in high-dimensional space, while the Monte Carlo sampling ensures robust statistics across a wide range of input distributions, making this a particularly valuable solution to analyze the impact of varying the top-$k$ in MoE models.

\paragraph{Stage 1: Per Layer MoE Top-K Perturbation Profiling:} Algorithm~\ref{alg:topk_perturbation} outlines our Monte Carlo-based method to evaluate the sensitivity 
% \SM{what do you mean by robustness? is this sensitivity?} yes sensitivity for random inputs. I will add this word here 
of each MoE layer under varying top-$k$ configurations. For each layer, we sample a random synthetic input tensor $\mathbf{X} \sim \mathcal{N}(0, 1)^{B \times L \times H}$ from the standard normal distribution. We first compute the baseline output using the default top-$k$ configuration, followed by computing outputs for each top-$k$ in the target search space. The perturbation induced by each top-$k$ is quantified using the Frobenius norm between the baseline output and the corresponding perturbed output. This process is repeated over millions of random input samples to obtain a statistically robust estimate of the average deviation for each candidate top-$k$. The Frobenius norm serves as a precise metric for capturing output magnitude shifts in high-dimensional space, while Monte Carlo sampling ensures diverse inputs. This sensitivity analysis provides a principled estimate of how active expert selection affects per-layer behavior. 
%\vspace{3mm}

\begin{algorithm}[H]
\caption{LExI Stage 1: Per Layer Top-$k$ Perturbation Loss Computation}
\label{alg:topk_perturbation}
\KwIn{
\begin{tabular}{ll}
    $\mathcal{M}_{\text{moe}}$: Mixture of Experts module (Gate $\mathcal{G}$ and Experts $\mathcal{E}$) & $T$: List of target top-$k$ values \\
    $k_{\text{base}}$: Baseline Pretrained top-$k$ & $B$: Batch size  \\
    $H$: Hidden Size & $L$: Sequence length \\
    $N_{\text{iter}}$: Number of iterations \\
    % $\mathcal{E}$: Experts module
    % $N_{\text{iter}}$: Number of Monte Carlo iterations
\end{tabular}
}
\KwOut{Average Frobenius Norm per top-$k$}

$\mathcal{D} \gets \{k: \emptyset\ \forall k \in T\}$\;

\For{$i \gets 1$ \textbf{to} $N_{\text{iter}}$}{
    
    Sample a input tensor from Normal Distribution: $\mathbf{X} \sim \mathcal{N}(0, 1)^{B \times L \times H}$\;
    
    \texttt{UpdateTopk}($\mathcal{M}_{\text{moe}}$, $k_{\text{base}}$)\;
    
    $\mathbf{Y}_{\text{base}} \gets f_{\text{moe}}(\mathbf{X})$ %\tcp*{Eq. (1) in paper}

    \ForEach{$k \in T$}{
        \texttt{UpdateTopk}($\mathcal{M}_{\text{moe}}$, $k$)\;
        $\mathbf{Y}_{\text{perturbed}}^{j} \gets f_{\text{moe}}(\mathbf{X})$\;

        $\Delta \gets \|\mathbf{Y}_{\text{perturbed}}^{j} - \mathbf{Y}_{\text{base}}\|_F$\;
        $\mathcal{D}[k] \gets \mathcal{D}[k] \cup \{\Delta\}$
    }
}

\ForEach{$k \in T$}{
    $\bar{\Delta}_k \gets \frac{1}{N_{\text{iter}}} \sum_{\delta \in \mathcal{D}[k]} \delta$\;
    $\mathcal{D}[k] \gets \bar{\Delta}_k$
}
\Return{$\mathcal{D}$}
\end{algorithm}
%\vspace{3mm}

% We observe different trends across different models. 

% The earlier layers Mixtral and MiniCPM are more robust to reduced experts (low $\Delta_k$), while later layers show larger deviations under top-$k$ reduction. 

% Contrary to the commonly observed trend that early layers are more sensitive to any architectural or precision perturbations (\cite{dong2019hawq}), our analysis on randomize input sampling reveals that initial layers in MoE models exhibit lower sensitivity to changes in the number of active experts compared to deeper layers. 

% Howver, in Qwen model, the initial layers are more sensitive to pertubing the top-$k$. DeepSeek models exhibit a bel curve where initial and last layers are more sensitive. 

\begin{figure*}[t]
    \centering

    \begin{subfigure}[b]{\textwidth}
        \centering
        \includegraphics[width=\linewidth, trim=0 0 0 0, clip]{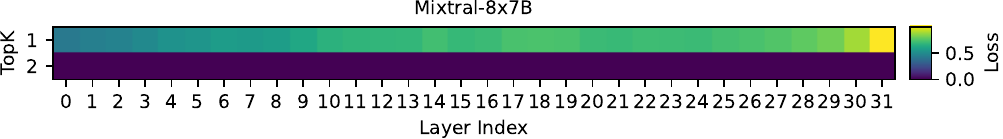}
    \end{subfigure}
    \vspace{-0.5ex}

    % \begin{subfigure}[b]{\textwidth}
    %     \centering
    %     \includegraphics[width=\linewidth, trim=0 0 0 0, clip]{heatmaps/MiniCPM_heatmap.pdf}
    % \end{subfigure}
    % \vspace{-0.5ex}

    \begin{subfigure}[b]{0.8\textwidth}
        \centering
        \includegraphics[width=\linewidth, trim=0 0 0 0, clip]{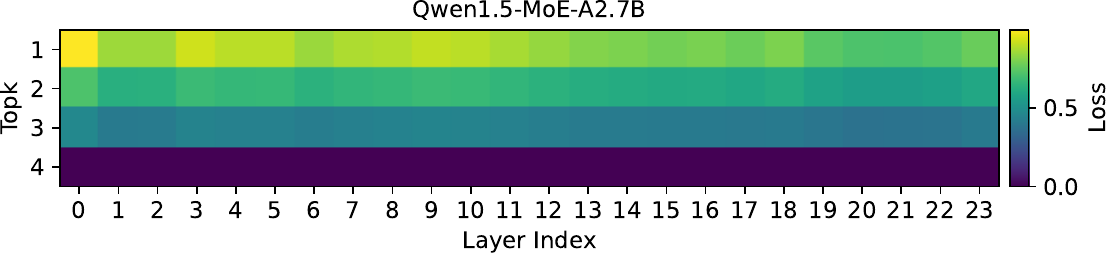}
    \end{subfigure}

    % \begin{subfigure}[b]{0.8\textwidth}
    %     \centering
    %     \includegraphics[width=\linewidth, trim=0 0 0 0, clip]{heatmaps/DeepSeekV2_heatmap.pdf}
    % \end{subfigure}
    % \vspace{-0.5ex}

\begin{minipage}[b]{0.49\textwidth}
    \centering
    \includegraphics[width=\linewidth, trim=0 0 0 0, clip]{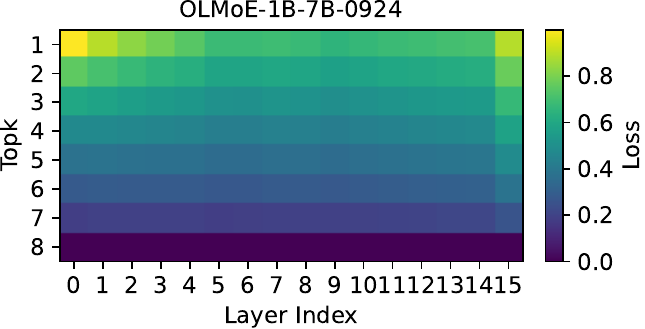}
\end{minipage}
\hfill
\begin{minipage}[b]{0.45\textwidth}
    \centering
    \includegraphics[width=\linewidth, trim=0 0 0 0, clip]{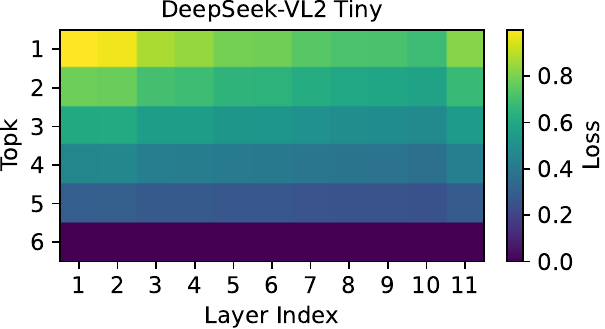}
\end{minipage}
\vspace{-0.5ex}

    \caption{Top-$k$ sensitivity analysis. The heatmap plots depict the layer-wise output deviation with respect to changing the top-$k$. The initial layers in Mixtral model are less sensitive to top-$k$ perturbation than deeper layers, while OLMoE exhibits a bell curve pattern where initial and last layers are more sensitive. Heatmaps for MiniCPM and DeepSeekV2 are shown in Appendix \ref{appendix:heatmap}. }
    \label{fig:topk_sensitivity}
\end{figure*}

\textit{Top-$k$ Perturbation Sensitivity Analysis:} Figure~\ref{fig:topk_sensitivity} visualizes the normalized sensitivity of different MoEs under various top-$k$s in the search space, measured using the Perturbation Loss ($\Delta_k$). Higher values of $\Delta_k$ indicate greater deviation from the baseline behavior, suggesting stronger sensitivity to changes in top-$k$. The sensitivity profiles vary notably across MoEs. For Mixtral-8x7B, early layers demonstrate greater sensitivity to reductions in the number of active experts, as indicated by lower perturbation loss, while later layers are more sensitive under top-$k$ perturbation. Interestingly, this finding diverges from prior works (\cite{dong2019hawq}), which suggests that early layers are typically more susceptible to architectural or precision perturbations. In contrast, Qwen1.5-MoE-A2.7B exhibits a reversed pattern where early layers are particularly sensitive to top-$k$ perturbations. DeepSeek model displays a bell-shaped sensitivity profile, with both initial and final layers demonstrating higher perturbation loss, while intermediate layers remain relatively stable. 
These findings have practical implications for adaptive expert selection strategies suggesting that latency can be can be optimized by reducing the number of active experts in more robust (low-sensitivity) layers.
% These insights are critical for designing adaptive expert selection policies tailored to different models, allocating more active experts in low-sensitivity layers. 

% \begin{figure*}
%     \centering
%      \includegraphics[width=0.9\linewidth]{heatmaps/Heatmap_all.pdf}
%         \caption{MoE}   
%         \vspace{-4mm}
%         \label{fig:frob_loss}
%     \captionsetup{justification=centering}
% \end{figure*}

% selection that operates on an allocation vector $k=(k_1,\dots,k_L)$ given input parameters ,  (minimum and maximum per-layer $k$), $B$ (total budget of active components), $L$ (number of layers), $N_{\text{pop}}$ (population size), $G_{\max}$ (max generations), and $\eta_{\text{mut}}$ (mutation rate). 

%\SM{Algorithm 2 doesnt seem to be referenced in the text}
\begin{algorithm}[H]
\caption{LExI Stage 2: Evolutionary Top-$k$ Allocation Optimization with Proxy}
\label{alg:evolutionary_topk}
\KwIn{
\begin{tabular}{ll}
    $\mathcal{D}$: TopK Perturbed Frobenius Norm Loss & $B$: Total Active Expert budget \\
    $k_{\min}$: Minimum topk per layer & $k_{\max}$: Maximum topk per layer \\
    $N_{\text{pop}}$: Population size & $G_{\text{max}}$: Maximum generations \\
    $\eta_{\text{mut}}$: Mutation rate & $L$: Number of Layers \\
\end{tabular}
}
\KwOut{Optimal topk allocation $\mathbf{k}^* = (k_1, ..., k_L)$}

Initialize population $\mathcal{P} \gets \{\mathbf{k}_i\}$ where $\mathbf{k}_i$ satisfies: \\
\quad $\sum_{j=1}^L k_j = B$ \textit{(Model budget constraint)} and
\quad $k_{\min}^j \leq k_j \leq k_{\max}^j\ \forall j$ \textit{(layer constraints)}

\For{$g \gets 1$ \textbf{to} $G_{\text{max}}$}{
    Evaluate fitness: $\phi(\mathbf{k}) = \sum_{j=1}^L \mathcal{D}_j(k_j)$ %\tcp*{Aggregate correlations}
    
    Select parents via tournament: $\mathbf{p}_1, \mathbf{p}_2 \gets \arg\min_{\mathbf{k} \in \mathcal{P}} \phi(\mathbf{k})$
    
    Generate offspring via \textsc{Crossover}: \\
    \quad $k'_j \gets \alpha_j p_{1,j} + (1-\alpha_j)p_{2,j}$ \quad $\alpha_j \sim \text{Bernoulli}(0.5)$
    
    Apply \textsc{Mutation}: \\
    \quad $k''_j \gets k'_j + \Delta_j$ \quad $\Delta_j \in \{-1,0,+1\}$ with $\sum_j \Delta_j = 0$
    
    Project to feasible space: \\
    \quad $\mathbf{k}''' \gets \text{Proj}(\mathbf{k}'')$ s.t. constraints hold
    
    Update population: \\
    \quad $\mathcal{P} \gets \mathcal{P} \cup \{\mathbf{k}'''\}$
}

\Return{$\mathbf{k}^* = \arg\min_{\mathbf{k} \in \mathcal{P}} \phi(\mathbf{k})$}
\end{algorithm}

\paragraph{Stage 2: Evolutionary Search with Proxy:}

LExI utilizes the proxies generated in the previous step to guide the evolutionary algorithm to allocate layer-wise top-$k$ as described in Algorithm \ref{alg:evolutionary_topk}. We define the total active expert budget (i.e. total number of active experts across all layers) $B$, minimum ($k_{\min}$) and maximum ($k_{\max}$) number of top-$k$ per layer. The objective is to find a feasible allocation $k^* = (k_1^,\dots,k_L)$ which minimizes the total layer-wise loss $\sum_{j=1}^L \mathcal{D}j(k_j)$ (the sum of TopK Perturbed Frobenius Norm losses) across $L$ layers, subject to the budget constraint $\sum{j=1}^L k_j = B$ and per-layer limits $k_{\min} \le k_j \le k_{\max}$ for all $j$. We initialize a population of $N_{\text{pop}}$ allocations (each satisfying the constraints) and then evolve this population over $G_{\max}$ generations. In each generation, every candidate solution is evaluated by its fitness $\phi(k) = \sum_{j=1}^L \mathcal{D}j(k_j)$, and parent solutions are chosen (e.g. via tournament selection) to produce offspring. A pair of parents is recombined using uniform crossover, wherein each layer’s allocation $k_j$ in the offspring is inherited from one of the two parents. The offspring is then mutated and ensured that the total budget $B$ remains unchanged (any increment in one layer’s $k_j$ is balanced by a decrement in another). After mutation, the new solution is added to the population. 
% to enforce $k_{\min} \le k_j \le k_{\max}$ and $\sum_j k_j = B$.
This evolutionary loop of selection, crossover, mutation, and repair is repeated for up to $G_{\max}$ generations, after which the best found allocation $k^*$ (the one minimizing $\phi(k)$) is returned. By maintaining the proxies, our LExI algorithm effectively navigates the combinatorial search space of discrete allocations and finds solutions fast without needing to load the actual model, making it well-suited for optimizing top-$k$ selection under various global active expert budgets where gradient-based methods require tremendous computational resources. 

% adding a small perturbation vector $\Delta = (\Delta_1,\dots,\Delta_L)$ with each $\Delta_j \in {-1,0,+1}$ and $\sum{j=1}^L \Delta_j = 0$, and 

% \subsection{Why it works?}

% \subsection{Searched Configurations}

%% file: Sections/5_results.tex
\section{Results and Evaluation}
\label{section:results}

This section presents our evaluation results and key insights from these runs. We compare our results with the baseline model, Inter Pruning (\cite{lu2024not}) and Intra Pruning (\cite{yang2024moe}).

\paragraph{LM-Eval Results. } 

Figure \ref{fig:lm_eval} presents a comprehensive comparison of average accuracy versus throughput across five MoE models on nine LM-Eval benchmarks. Traditional pruning methods, inter-expert (red) and intra-expert (blue), consistently demonstrate a trade-off:  reducing the number of parameters improves throughput but significantly degrades accuracy. In contrast, our proposed method, LExI (green), achieves a more favorable accuracy-throughput balance across all models. 
On OLMoE-1B-7B (Figure \ref{fig:lm_eval}a), LExI with the active expert budget ($B$) = 100
% \SM{there are three points for Lexi in 4(a)}
achieves the same throughput as $50\%$ intra-pruning while delivering $+10\%$ higher accuracy, and outperforms the $50\%$ inter-pruning baseline by +15\% higher accuracy. On Qwen1.5-MoE, LExI offers at least $+5.1\%$ higher throughput compared to both inter and intra pruning, with a consistent $+0.5$ \% accuracy gain. For MiniCPM-MoE, LExI achieves $+15\%$ higher accuracy than $25\%$ inter-pruned models at equivalent throughput, demonstrating superior accuracy-throughput tradeoffs. For Mixtral-8x7B, LExI surpasses the inter-pruned baseline by $+10\%$ accuracy at nearly identical throughput. Finally, on DeepSeekV2-Lite, LExI outperforms inter pruning with $+6.5\%$ higher throughput at equal accuracy, and recovers $+6$ accuracy points over intra pruning with only a minor throughput compromise.
Notably, in Qwen1.5-MoE and MiniCPM-MoE, LExI not only avoids accuracy degradation but also achieves throughput comparable to or better than 50\% inter- or intra-expert pruning.
In summary, \emph{LExI preserves accuracy close to the base model while delivering substantial throughput gains, often surpassing both pruning approaches}. 
% For example, in OLMoE-1B-7B, Mixtral-8x7B, and DeepSeekV2-Lite, LExI maintains accuracy within close proximity to the base model while offering notable throughput gains, often outperforming both pruning strategies. 
These results highlight the robustness and effectiveness of LExI’s expert budget reallocation in preserving model performance while improving inference efficiency, demonstrating its superiority over uniform pruning.

\input{Sections/5_results_lm_eval}

\noindent\textbf{Long context Evaluation. } Figure \ref{fig:qasper_longbench} presents an accuracy-throughput comparison on the Qasper dataset for three different MoE models. Inter and intra pruning methods consistently reduce throughput but often come at a sharp cost in F1 score, particularly under higher pruning ratios. In contrast, our LExI method achieves a more favorable trade-off by dynamically reducing the number of active experts per layer based on sensitivity, leading to Pareto 
% \SM{of what quantities?}
improvements (F1 score and throughput) in multiple cases. For example, in Qwen1.5 and DeepSeek models, LExI achieves higher throughput than the base model while maintaining competitive or superior F1 scores, demonstrating its efficacy in preserving task accuracy while enhancing inference efficiency. On Qwen1.5-MoE-A2.7B, LExI achieves a score of $35.5$ at $\sim4.1\text{k}$ throughput, outperforming inter pruning (F1 $34$ at $\sim3.9\text{k}$) and intra pruning (F1 $30$ at $\sim3.75\text{k}$) with a $+0.5$--$5.5$ F1 gain and $+5.1\%$--$9.3\%$ higher throughput.

\input{Sections/5_results_LongBench}

\noindent\textbf{Passkey Retrieval Task. } Figure \ref{fig:passkey} illustrates the trade-off between throughput and average accuracy for the Passkey Retrieval task across five MoE models. This task evaluates a model's ability to extract precise key information embedded in distractive or noisy contexts, demanding both precision and robustness. Traditional inter- and intra-pruning approaches generally degrade performance, with noticeable accuracy drops even at moderate pruning levels. In contrast, our proposed LExI method consistently outperforms these baselines by achieving higher or comparable accuracy while significantly improving throughput. Notably, in models like OLMoE-1B-7B and Qwen1.5-MoE, LExI nearly restores or surpasses the base model’s accuracy while offering improved efficiency. These results highlight LExI’s strength in preserving critical retrieval capabilities under expert budget constraints, making it a superior choice for precision-sensitive tasks like passkey extraction.

\input{Sections/5_Results_PassKey}

\noindent\textbf{Perplexity Evaluation. }
Figure \ref{fig:perplexity} presents a comparison of perplexity vs. throughput across multiple MoE models and datasets, highlighting the limitations of traditional pruning techniques. Across models, LExI consistently offers a better accuracy-efficiency benefit, outperforming both the pruning baselines. Notably, pruning methods often yield modest throughput gains but at the cost of substantial perplexity degradation, especially evident in the OLMoE and Mixtral cases. In contrast, LExI achieves throughput improvements close to aggressive pruning levels while almost preserving the perplexity of the base model. For example, on Mixtral-8x7B, LExI achieves $\sim$2.4k throughput at $\sim$23 perplexity on C4 dataset and $\sim$2.2k throughput at $\sim$10 perplexity on WikiText, whereas inter pruning attains the same speedup with double the perplexity. This indicates that static expert reduction via LExI enables smarter compute allocation, unlike pruning, which disrupts the expert-token mapping and leads to suboptimal routing and load imbalance.

% Figure \ref{fig:perplexity} presents a comparison of perplexity vs. throughput across multiple MoE models and datasets, highlighting the limitations of traditional pruning techniques and the efficacy of our proposed LExI method. Across all models, LExI consistently offers a better accuracy-efficiency benefit, outperforming both the pruning baselines. Notably, pruning methods often yield modest throughput gains but at the cost of substantial perplexity degradation, especially evident in the OLMoE and Mixtral cases. In contrast, LExI achieves throughput improvements close to aggressive pruning levels while almost preserving the perplexity of the base model. This indicates that static expert reduction via LExI enables smarter compute allocation, unlike pruning, which disrupts the expert-token mapping and leads to suboptimal routing and load imbalance.

\input{Sections/5_results_perplexity}

\noindent\textbf{Ablation Study on Vision Language Domain. }
Figure \ref{fig:vlmevalkit} evaluates the LExI method on DeepSeekVL2-Tiny across four vision-language tasks. While intra-pruning peaks at 25\% intra pruning at the cost of sharp accuracy drops, its performance is highly unstable and inconsistent across tasks. Inter-pruning, on the other hand, exhibits a flat trade-off curve that fails to provide significant speedups. In contrast, LExI consistently achieves superior accuracy and throughput balance, yielding improvements without compromising performance. Unlike pruning, which disrupts expert specialization and introduces fragility, LExI leverages structured top-$k$ expert budget selection that preserves model capacity while enabling efficient routing, making it a more robust alternative for optimizing sparse VLMs.

\input{Sections/5_results_vlmeval_kit}

%% file: Sections/5_results_lm_eval.tex
\begin{figure*}[t]
    \centering
    % Legend spanning full width at the top
    \includegraphics[width=\textwidth]{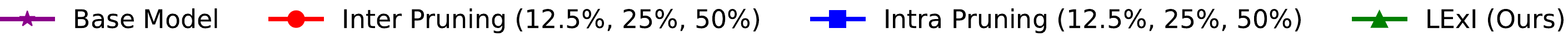}\\[-0.3ex]

    % Five plots arranged horizontally
    \begin{subfigure}[b]{0.195\textwidth}
        \centering
        \includegraphics[width=\linewidth]{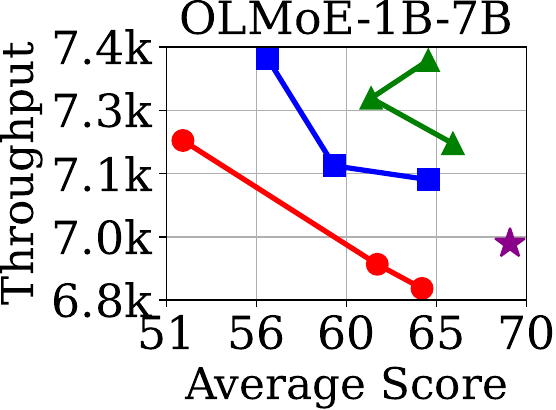}
        
        % \caption{\scriptsize\centering OlmoE-1B-7B-Instruct \protect\\ (B = 64, 72, 100)}
        \caption{\scriptsize \shortstack{OlmoE-1B-7B-Instruct \\ (B = 64, 72, 100)}}

    \end{subfigure}\hspace{0pt}
    \begin{subfigure}[b]{0.195\textwidth}
        \centering
        \includegraphics[width=\linewidth]{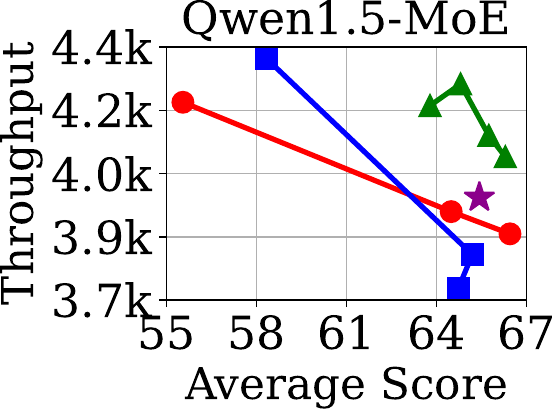}
        
        % \caption{\scriptsize\centering Qwen1.5-MoE-A2.7B \protect\\ (B = 48, 56, 72, 80)}
        \caption{\scriptsize \shortstack{Qwen1.5-MoE-A2.7B \\ (B = 48, 56, 72, 80)}}
        
    \end{subfigure}\hspace{0pt}
    \begin{subfigure}[b]{0.195\textwidth}
        \centering
        \includegraphics[width=\linewidth]{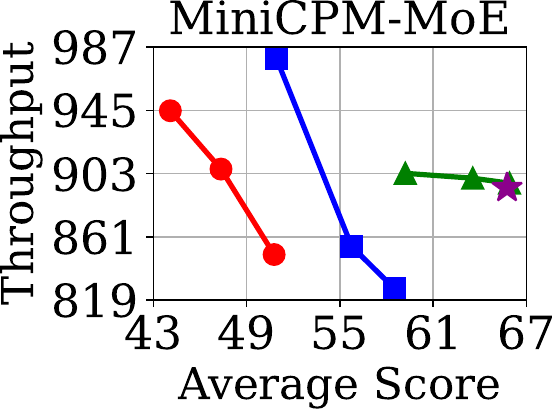}
         
         % \caption{\scriptsize\centering MiniCPM-MoE-8x2B \protect\\ (B = 50, 60, 70)}
        \caption{\scriptsize \shortstack{MiniCPM-MoE-8x2B \\ (B = 50, 60, 70)}}
         
    \end{subfigure}\hspace{0pt}
    \begin{subfigure}[b]{0.195\textwidth}
        \centering
        \includegraphics[width=\linewidth]{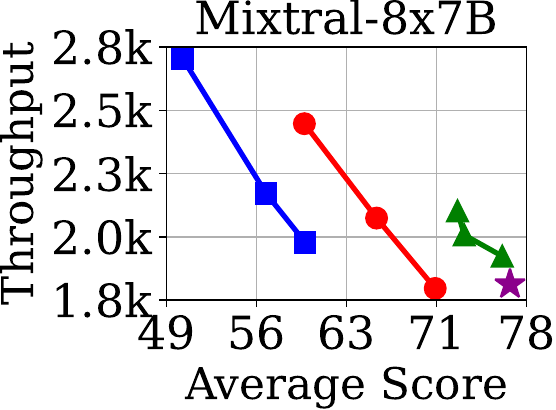}
        
        % \caption{\scriptsize\centering Mixtral-8x7B-Instruct \protect\\ (B = 40, 48, 56)}
        \caption{\scriptsize \shortstack{Mixtral-8x7B-Instruct \\ (B = 40, 48, 56)}}
        
    \end{subfigure}\hspace{0pt}
    \begin{subfigure}[b]{0.195\textwidth}
        \centering
        \includegraphics[width=\linewidth]{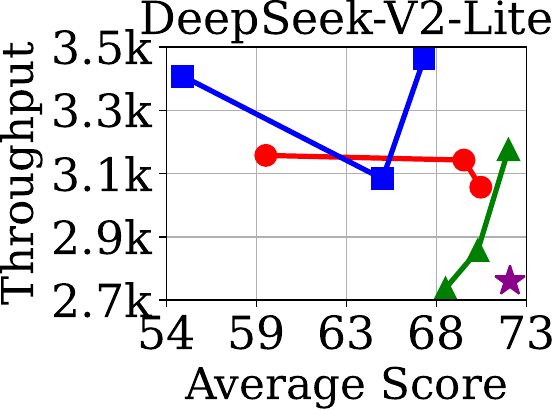}
         % \caption{\scriptsize\centering DeepSeekV2-Lite-Chat \protect\\ (B = 78, 104, 130)}
         \caption{\scriptsize \shortstack{DeepSeekV2-Lite-Chat \\ (B = 78, 104, 130)}}
         
    \end{subfigure}

    \caption{Average Accuracy ($\uparrow$) vs Throughput ($\uparrow$) on 9 LM-Eval Tasks (ARC-c, ARC-e, BoolQ, HellaSwag, MMLU, OBQA, RTE, WinoGrande). $B$: Active Expert Budget
    % \SM{what do the three points for LeXi correspond to?}
    }
    \label{fig:lm_eval}
\end{figure*}

%% file: Sections/5_results_LongBench.tex
\begin{figure*}[t]
    \centering
    \includegraphics[width=\textwidth]{Figures/LM_Eval/LExI_Legend.pdf}\\[-0.3ex]

    \begin{subfigure}[b]{0.21\textwidth}
        \centering
        \includegraphics[width=\linewidth]{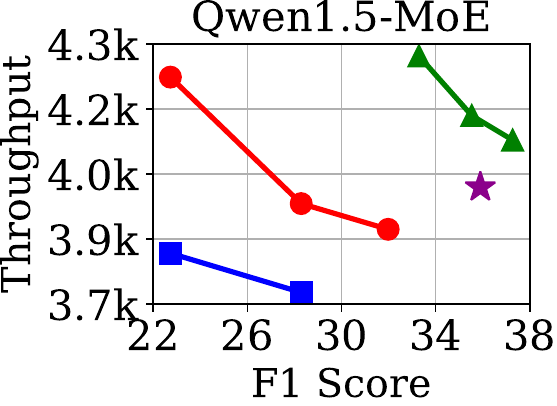}
         
         % \caption{\scriptsize\centering Qwen1.5-MoE-A2.7B \protect\\ (B = 48, 56, 72, 80)}
         \caption{\scriptsize \shortstack{Qwen1.5-MoE-A2.7B \\ (B = 48, 56, 72, 80)}}

    \end{subfigure}\hspace{0pt}
    \begin{subfigure}[b]{0.21\textwidth}
        \centering
        \includegraphics[width=\linewidth]{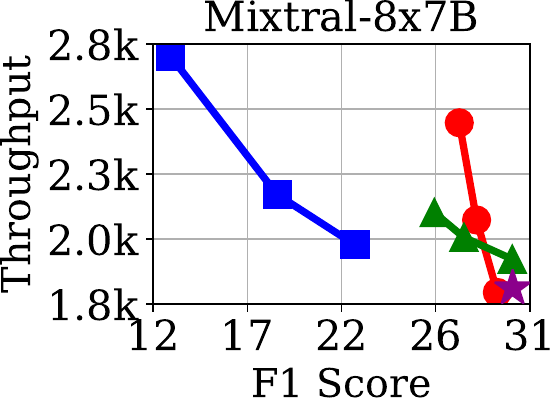}
         % \caption{\scriptsize\centering Mixtral-8x7B-Instruct \protect\\ (B = 40, 48, 56)}
         \caption{\scriptsize \shortstack{Mixtral-8x7B-Instruct \\ (B = 40, 48, 56)}}

    \end{subfigure}\hspace{0pt}
    \begin{subfigure}[b]{0.21\textwidth}
        \centering
        \includegraphics[width=\linewidth]{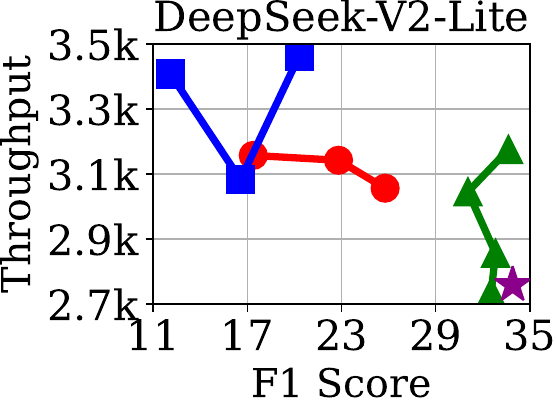}
         % \caption{\scriptsize\centering DeepSeekV2-Lite-Chat \protect\\ (B = 78, 104, 130)}
         \caption{\scriptsize \shortstack{DeepSeekV2-Lite-Chat \\ (B = 78, 104, 130)}}

    \end{subfigure}
    \caption{F1 Score ($\uparrow$) vs Throughput ($\uparrow$) on Qasper Dataset in LongBench. $B$: Active Expert Budget}
    \label{fig:qasper_longbench}
\end{figure*}

%% file: Sections/5_Results_PassKey.tex
\begin{figure*}[t]
    \centering
    \includegraphics[width=\textwidth]{Figures/LM_Eval/LExI_Legend.pdf}\\[0ex]

    \begin{subfigure}[b]{0.195\textwidth}
        \centering
        \includegraphics[width=\linewidth]{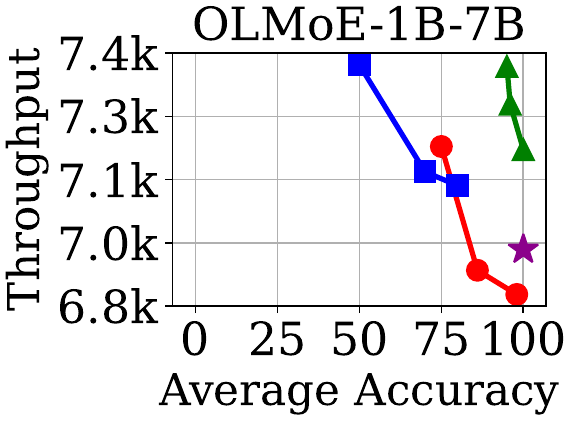}
        
        % \caption{\scriptsize\centering OlmoE-1B-7B-Instruct \protect\\ (B = 64, 72, 100)}
        \caption{\scriptsize \shortstack{OlmoE-1B-7B-Instruct \\ (B = 64, 72, 100)}}

    \end{subfigure}\hspace{0pt}
    \begin{subfigure}[b]{0.195\textwidth}
        \centering
        \includegraphics[width=\linewidth]{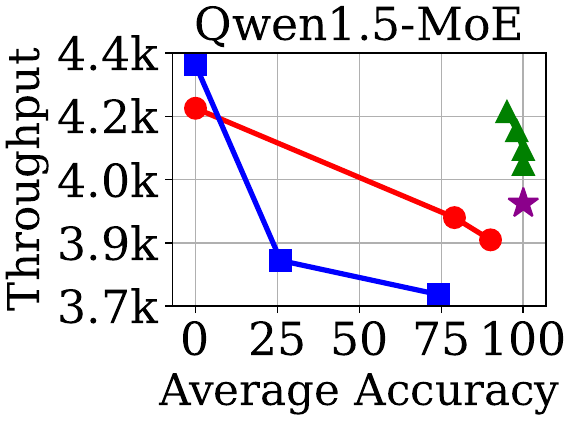}
        
        % \caption{\scriptsize\centering Qwen1.5-MoE-A2.7B \protect\\ (B = 48, 56, 72, 80)}
        \caption{\scriptsize \shortstack{Qwen1.5-MoE-A2.7B \\ (B = 48, 56, 72, 80)}}

    \end{subfigure}\hspace{0pt}
    \begin{subfigure}[b]{0.195\textwidth}
        \centering
        \includegraphics[width=\linewidth]{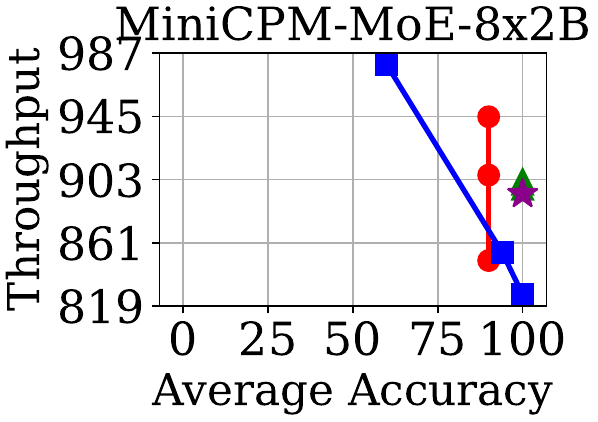}
         % \caption{\scriptsize\centering MiniCPM-MoE-8x2B \protect\\ (B = 50, 60, 70)}
         \caption{\scriptsize \shortstack{MiniCPM-MoE-8x2B \\ (B = 50, 60, 70)}}

    \end{subfigure}\hspace{0pt}
    \begin{subfigure}[b]{0.195\textwidth}
        \centering
        \includegraphics[width=\linewidth]{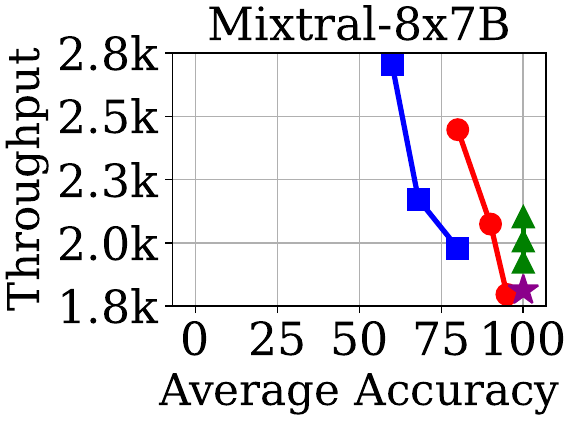}
        % \caption{\scriptsize\centering Mixtral-8x7B-Instruct \protect\\ (B = 40, 48, 56)}
        \caption{\scriptsize \shortstack{Mixtral-8x7B-Instruct \\ (B = 40, 48, 56)}}

    \end{subfigure}\hspace{0pt}
    \begin{subfigure}[b]{0.195\textwidth}
        \centering
        \includegraphics[width=\linewidth]{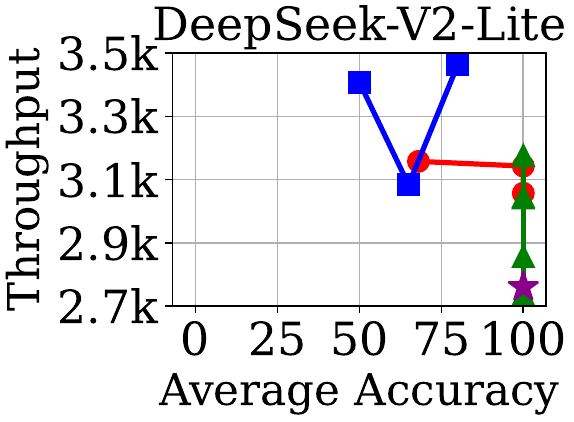}
        % \caption{\scriptsize\centering DeepSeekV2-Lite-Chat \protect\\ (B = 78, 104, 130)}
        \caption{\scriptsize \shortstack{DeepSeekV2-Lite-Chat \\ (B = 78, 104, 130)}}

    \end{subfigure}

    \caption{Passkey Retrieval Average ($\uparrow$) vs Throughput ($\uparrow$) Comparison. $B$: Active Expert Budget}
    \label{fig:passkey}
\end{figure*}

%% file: Sections/5_results_perplexity.tex
\begin{figure*}[t]
    \centering
    \includegraphics[width=\textwidth]{Figures/LM_Eval/LExI_Legend.pdf}\\[-0.3ex]
    \begin{subfigure}[b]{0.22\textwidth}
        \centering
        \includegraphics[width=\linewidth]{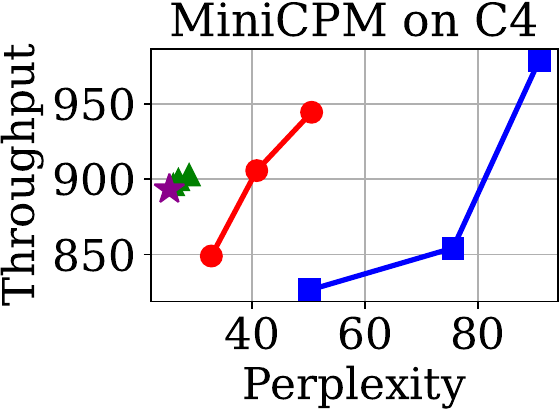}
        % \caption{\scriptsize\centering MiniCPM-MoE-8x2B \protect\\ (B = 50, 60, 70)}
        % \caption{\scriptsize \shortstack{MiniCPM-MoE-8x2B \\ (B = 50, 60, 70)}}
        \caption{\scriptsize \shortstack{MiniCPM-MoE-8x2B \\ (B = 50, 60, 70)}}
        % \caption{\scriptsize MiniCPM-MoE-8x2B \protect\linebreak (B = 50, 60, 70)}        
    \end{subfigure}\hspace{0pt}
    \begin{subfigure}[b]{0.22\textwidth}
        \centering
        \includegraphics[width=\linewidth]{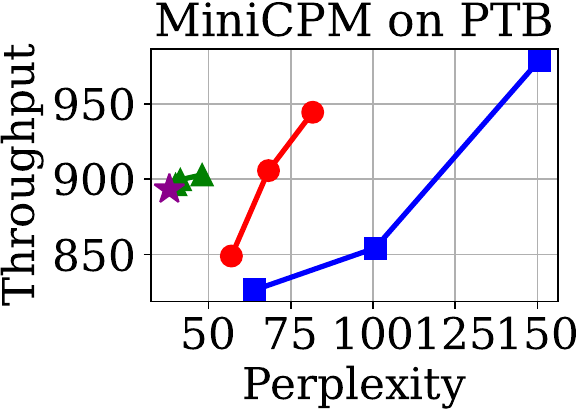}
         % \caption{\scriptsize\centering MiniCPM-MoE-8x2B \protect\\ (B = 50, 60, 70)}
         \caption{\scriptsize \shortstack{MiniCPM-MoE-8x2B \\ (B = 50, 60, 70)}}
    \end{subfigure}\hspace{0pt}
    \begin{subfigure}[b]{0.22\textwidth}
        \centering
        \includegraphics[width=\linewidth]{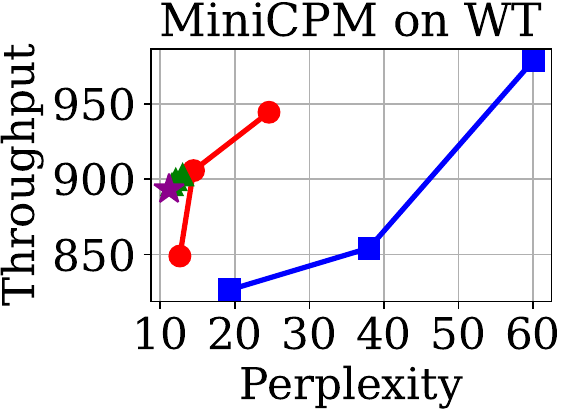}
         % \caption{\scriptsize\centering MiniCPM-MoE-8x2B \protect\\ (B = 50, 60, 70)}
         \caption{\scriptsize \shortstack{MiniCPM-MoE-8x2B \\ (B = 50, 60, 70)}}
    \end{subfigure}\hspace{0pt}
    \begin{subfigure}[b]{0.22\textwidth}
        \centering
        \includegraphics[width=\linewidth]{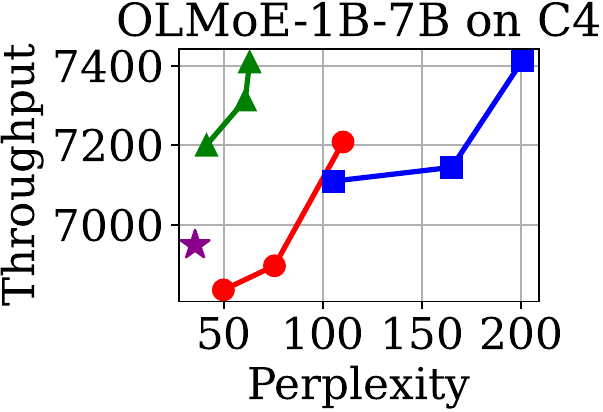}
         % \caption{\scriptsize\centering OlmoE-1B-7B-Instruct \protect\\ (B = 64, 72, 100)}
         % \caption{\scriptsize \shortstack{MiniCPM-MoE-8x2B \\ (B = 50, 60, 70)}}
         \caption{\scriptsize \shortstack{MiniCPM-MoE-8x2B \\ (B = 50, 60, 70)}}
    \end{subfigure}
    \begin{subfigure}[b]{0.22\textwidth}
        \centering
        \includegraphics[width=\linewidth]{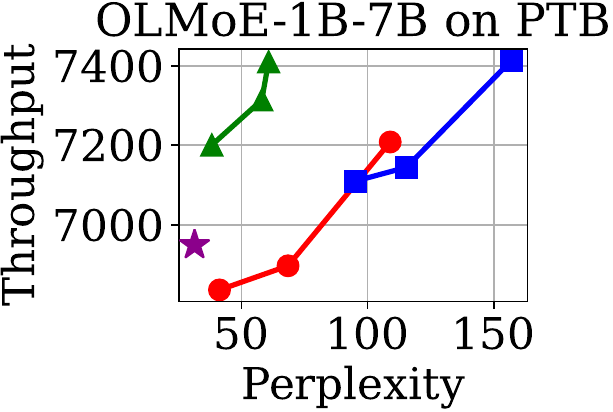}
        % \caption{\scriptsize\centering OlmoE-1B-7B-Instruct \protect\\ (B = 64, 72, 100)}
        \caption{\scriptsize \shortstack{OlmoE-1B-7B-Instruct \\ (B = 64, 72, 100)}}
    \end{subfigure}\hspace{0pt}
    \begin{subfigure}[b]{0.22\textwidth}
        \centering
        \includegraphics[width=\linewidth]{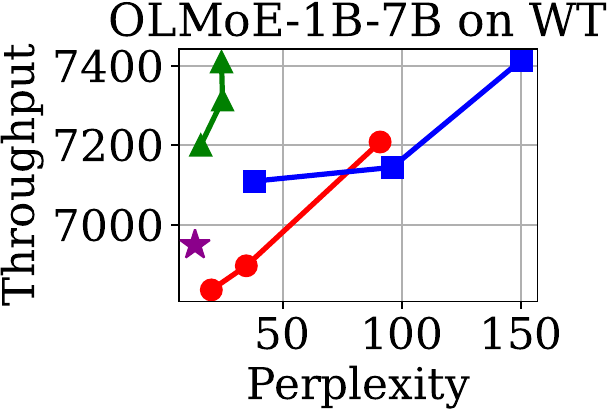}
         % \caption{\scriptsize\centering OlmoE-1B-7B-Instruct \protect\\ (B = 64, 72, 100)}
         \caption{\scriptsize \shortstack{OlmoE-1B-7B-Instruct \\ (B = 64, 72, 100)}}
    \end{subfigure}\hspace{0pt}
    \begin{subfigure}[b]{0.22\textwidth}
        \centering
        \includegraphics[width=\linewidth]{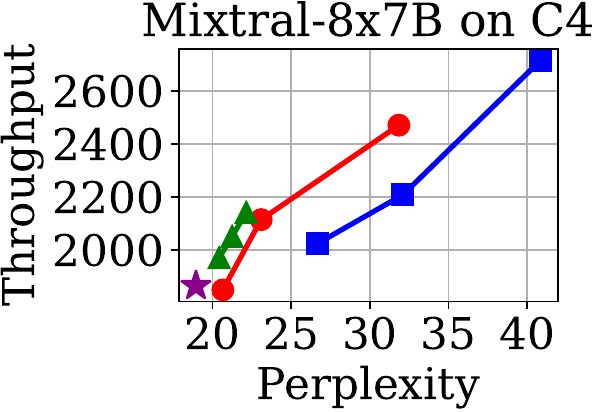}
         % \caption{\scriptsize\centering Mixtral-8x7B-Instruct \protect\\ (B = 40, 48, 56)}
         \caption{\scriptsize \shortstack{Mixtral-8x7B-Instruct \\ (B = 40, 48, 56)}}
    \end{subfigure}\hspace{0pt}
    \begin{subfigure}[b]{0.22\textwidth}
        \centering
        \includegraphics[width=\linewidth]{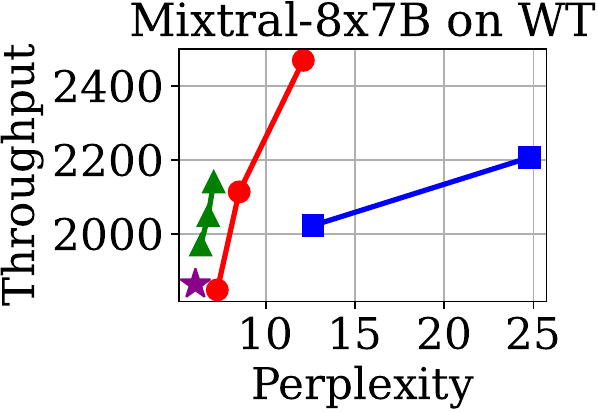}
        % \caption{\scriptsize\centering Mixtral-8x7B-Instruct \protect\\ (B = 40, 48, 56)}
        \caption{\scriptsize \shortstack{Mixtral-8x7B-Instruct \\ (B = 40, 48, 56)}}
    \end{subfigure}
    \caption{Perplexity ($\downarrow$) vs Throughput ($\uparrow$) on C4, PTB \& WikiText(WT) $B$: Active Expert Budget}
    \label{fig:perplexity}
\end{figure*}

%% file: Sections/5_results_vlmeval_kit.tex
\begin{figure*}[t]
    \centering
    \includegraphics[width=\textwidth]{Figures/LM_Eval/LExI_Legend.pdf}\\[-0.3ex]

    \begin{subfigure}[b]{0.22\textwidth}
        \centering
        \includegraphics[width=\linewidth]{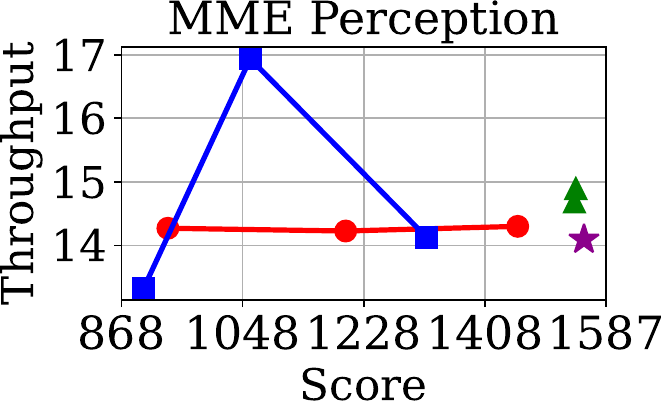}
         % \caption{\scriptsize\centering DeepSeekVL2-Tiny \protect\\ (B = 44, 48, 55)}
         \caption{\scriptsize \shortstack{DeepSeekVL2-Tiny \\ (B = 44, 48, 55)}}
    \end{subfigure}\hspace{0pt}
    \begin{subfigure}[b]{0.22\textwidth}
        \centering
        \includegraphics[width=\linewidth]{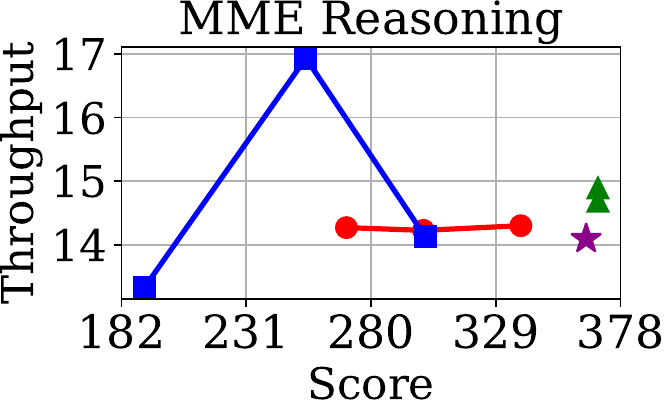}
        % \caption{\scriptsize\centering DeepSeekVL2-Tiny \protect\\ (B = 44, 48, 55)}
        % \caption{\scriptsize \shortstack{MiniCPM-MoE-8x2B \\ (B = 50, 60, 70)}}
        \caption{\scriptsize \shortstack{DeepSeekVL2-Tiny \\ (B = 44, 48, 55)}}
    \end{subfigure}\hspace{0pt}
    \begin{subfigure}[b]{0.22\textwidth}
        \centering
        \includegraphics[width=\linewidth]{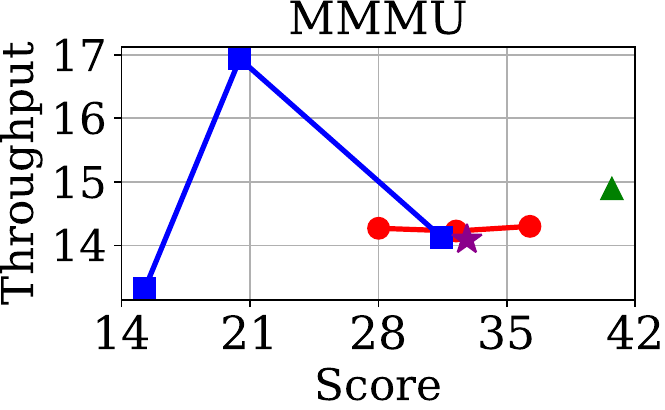}
         % \caption{\scriptsize\centering DeepSeekVL2-Tiny \protect\\ (B = 44, 48, 55)}
         % \caption{\scriptsize \shortstack{MiniCPM-MoE-8x2B \\ (B = 50, 60, 70)}}
         \caption{\scriptsize \shortstack{DeepSeekVL2-Tiny \\ (B = 44, 48, 55)}}
    \end{subfigure}\hspace{0pt}
    \begin{subfigure}[b]{0.22\textwidth}
        \centering
        \includegraphics[width=\linewidth]{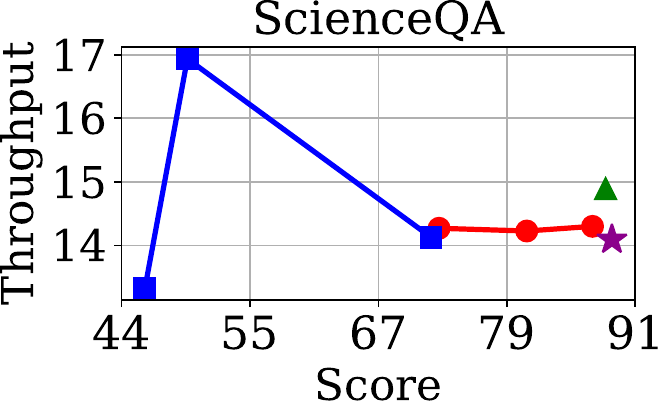}
        % \caption{\scriptsize\centering DeepSeekVL2-Tiny \protect\\ (B = 44, 48, 55)}
        % \caption{\scriptsize \shortstack{MiniCPM-MoE-8x2B \\ (B = 50, 60, 70)}}
        \caption{\scriptsize \shortstack{DeepSeekVL2-Tiny \\ (B = 44, 48, 55)}}
    \end{subfigure}
    \caption{DeepSeekVL2-Tiny: Average Accuracy ($\uparrow$) vs Throughput ($\uparrow$) on MME, MMMU, ScienceQA Tasks in VLMEvalKit. $B$: Active Expert Budget}
    \label{fig:vlmevalkit}
\end{figure*}

%% file: Sections/6_abalation.tex
% \section{Ablation Studies}

% \subsection{Random Search vs Evolutionary NAS}

% \subsection{Varying total Topk}

% \subsection{Mixture of Expert Heads}

%% file: Sections/7_conclusion.tex
\section{Limitations}
\label{sec:limitations}

% Our approach has two primary limitations. First, it does not reduce the memory footprint of the MoE model. Unlike prior expert pruning methods that explicitly target improving memory efficiency by removing parameters, our method focuses solely on optimizing computational performance during inference. As a result, it is less effective in memory-constrained deployment scenarios. However, our method can effectively be combined with existing MoE pruning methods to optimize both compute and memory. Second, the method may underperform in settings where the top-$k$ expert search space is inherently limited. For instance, in models like Llama-4, which are pretrained with a single active expert per layer, our strategy has no room to reduce the number of active experts further, rendering it inapplicable.

Our approach has two primary limitations. First, it does not reduce the memory footprint of the MoE model. Unlike prior expert pruning methods that explicitly target improving memory efficiency by removing parameters, our method focuses solely on optimizing computational performance during inference. This means that while our approach can significantly speed up inference by reducing the number of expert computations, it does not reduce model size. As a result, it is less effective in memory-constrained deployment scenarios.  
% As such, in environments with strict memory constraints—such as mobile or edge devices—our method offers limited benefit on its own. 
Nevertheless, our method can be effectively combined with existing MoE pruning methods, enabling a joint optimization of both computational efficiency and model memory. Second, our method may underperform in settings where the top-\$k\$ expert search space is inherently limited. Our approach relies on selectively reducing the number of active experts during inference to gain computational efficiency. However, in architectures such as Llama-4, where each MoE layer is pretrained with only a single active expert, there is no flexibility to reduce active experts further. In such scenarios, our method becomes inapplicable. 

% , as the opportunity for selective computation is already exhausted. 

% This highlights a key boundary of our technique—it is best suited for models that use a larger top-\$k\$ routing during pretraining, allowing room for optimization at inference time.

\section{Conclusion}

In this paper, we propose LExI, a framework designed to determine the optimal number of active experts per layer of a pretrained MoE model. LExI achieves better throughput than both inter and intra expert pruning baseline methods across six different MoE models. Unlike uniform expert pruning, which can yield speedups only at the cost of substantial accuracy loss, our method delivers substantial throughput gains while preserving model accuracy/perplexity close to the original baseline. For example, on OLMoE-1B-7B, Mixtral-8x7B, and DeepSeek, LExI maintains accuracy nearly identical to the unpruned model yet attains notably higher throughput, often surpassing the accuracy of pruned models. In certain scenarios, this layer-adaptive approach achieves  higher throughput than the base model with the same task performance, highlighting the robustness of the proposed approach. For example, on OLMoE-1B-7B model, our approach matches the throughput of 50\% Intra pruned model with 10\% better accuracy. Also, LExI’s benefits come with no retraining or calibration dataset requirements. It is a data-free post-training method that optimizes without any calibration dataset or fine-tuning. 
%This makes LExI a practical inference run time optimization technique, which by determining the optimal number of active experts per layer, achieves lower inference latency, minimizes inter-GPU communication overhead, and memory bandwidth usage, while incurring negligible accuracy drop.
This makes LExI a practical inference-time optimization technique that reduces latency, inter-GPU communication overhead, and memory bandwidth usage by determining the optimal number of active experts per layer, all while incurring negligible accuracy loss.

%% file: Sections/8_Appendix.tex
\appendix

% \section{Appendix / supplemental material}

\section{Appendix}

\subsection{Mixture of Experts Model Setup} \label{appendix:experimental_llms_vlms}

Table \ref{fig:LLMs_VLMs} illustrates the hyperparamters of each MoE model we utilized in our evaluation. 

\begin{table}[H]
\centering
\caption{LLM and VLM MoE Models}
\begin{tabular}{|c|c|c|c|c|c|}
\hline
% \textbf{Model Type} & 
\textbf{Model} & \textbf{\#P (B)} & \textbf{\#Layers} & \textbf{\#Experts} & \textbf{TopK} & \textbf{FFN Dim} \\
\hline
% \multirow{6}{*}{LLM} 
DeepSeek VL2-Tiny      & 3     & 12 & 64  & 6 & 896\\
OLMoE-1B-7B-0125-Instruct & 6.92  & 16 & 64   & 8 & 1024\\
Qwen1.5-MoE-A2.7B-Chat      & 14.3  & 24 & 60  & 4 & 1408\\
DeepSeek-V2-Lite-Chat            & 15.7  & 27 & 64  & 6 & 1408\\
MiniCPM-MoE-8x2B  & 17  & 40 & 8  & 2 & 5760\\
Mixtral-8x7B-Instruct-v0.1 & 46.7  & 32 & 8   & 2 & 14336\\
  
  % & AI21-Jamba-Mini-1.6 & 51.6  & 32 & 16 & 2 & 14336\\
% \hline
% \multirow{1}{*}{VLM} 
  % & 
  % & MolmoE-1B              & 7.2   & 16 & 64  & 8 & --\\
  % & DeepSeek VL2-Small     & 16    & 27 & 64  & 6 & --\\
  % & DeepSeek VL2           & 27    & 30 & 72  & 6 & --\\
\hline
\end{tabular}
\label{fig:LLMs_VLMs}
\end{table}

\subsection{Additional Heatmaps for top-K sensitivity}
\label{appendix:heatmap}

\begin{figure*}[!ht]
    \centering

    \begin{subfigure}[b]{\textwidth}
        \centering
        \includegraphics[width=\linewidth, trim=0 0 0 0, clip]{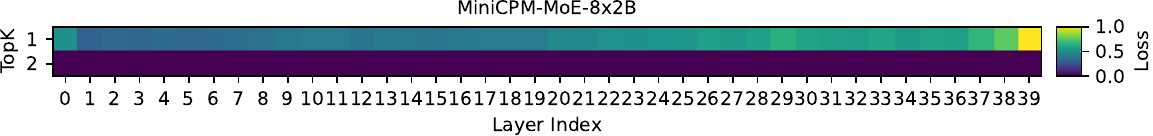}
    \end{subfigure}
    \vspace{-0.5ex}

    \begin{subfigure}[b]{0.8\textwidth}
        \centering
        \includegraphics[width=\linewidth, trim=0 0 0 0, clip]{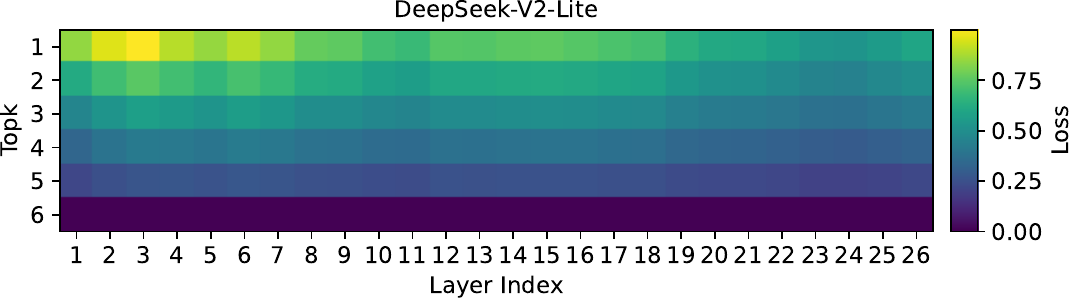}
    \end{subfigure}
    \vspace{-0.5ex}

% \begin{minipage}[b]{0.49\textwidth}
%     \centering
%     \includegraphics[width=\linewidth, trim=0 0 0 0, clip]{heatmaps/OLMoE_1B_heatmap.pdf}
% \end{minipage}
% \hfill
% \begin{minipage}[b]{0.45\textwidth}
%     \centering
%     \includegraphics[width=\linewidth, trim=0 0 0 0, clip]{heatmaps/deepseek_vl2_tiny_heatmap.pdf}
% \end{minipage}
% \vspace{-0.5ex}

    \caption{Top-$k$ sensitivity analysis across MiniCPM-MoE-8x2B and DeepSeekV2-Lite. The plots depict the layer-wise output deviation with respect to changing the top-$k$. The initial layers in MiniCPM model are less sensitive to topk perturbation than deeper layers, while DeepSeekV2 exhibits a bell curve pattern where initial and last layers are more sensitive.}
    \label{fig:topk_sensitivity_appendix}
\end{figure*}

Figure \ref{fig:topk_sensitivity_appendix} illustrates the topk sensitivity heatmaps for MiniCPM-MoE and DeepSeekV2 Lite Chat model based on Algorithm \ref{alg:topk_perturbation}. 

